\documentclass{article}
\usepackage{ijcai25}

\usepackage{times}
\usepackage{soul}
\usepackage{url}
\usepackage[hidelinks]{hyperref}
\usepackage[utf8]{inputenc}
\usepackage[small]{caption}
\usepackage{graphicx}
\usepackage{amsmath}
\usepackage{amsthm}
\usepackage{booktabs}
\usepackage{algorithm}
\usepackage{algorithmic}
\usepackage[switch]{lineno}
\usepackage{multirow}
\usepackage{makecell}
\usepackage{array}
\usepackage{url}
\usepackage{color}
\usepackage{caption}
\usepackage{subcaption}
\usepackage{amssymb}
\usepackage{xr-hyper}
\usepackage[dvipsnames]{xcolor}
\definecolor{mypink}{RGB}{227, 119, 194}

\title{MQADet: A Plug-and-Play Paradigm for Enhancing Open-Vocabulary Object Detection via Multimodal Question Answering}

\author{
Caixiong Li$^{1,4*}$ \and
Xiongwei Zhao$^2$\footnote{Equal contribution} \and
Jinhang Zhang$^3$ \and
Xing Zhang$^{1,4}$\footnote{Corresponding author} \and Qihao Sun$^3$ \and
Zhou Wu$^{5}$ \\
\affiliations
$^1$School of Computer and Information Science, Qinghai Institute of Technology\\
$^2$School of Information Science and Technology, Harbin Institute of Technology (Shen Zhen)\\
$^3$The State Key Laboratory of Robotics and System, Harbin Institute of Technology\\
$^4$School of Computer Science and Technology, Qinghai University \\
$^5$Eryuan Digital Technology Co., Ltd.\\
\emails
cxli@qhit.edu.cn,
xwzhao@stu.hit.edu.cn,
24b908088@stu.hit.edu.cn,
xzhang@qhit.edu.cn,
23S008047@stu.hit.edu.cn,
chouwu8sone@gmail.com
}

\begin{document}

\maketitle

\begin{abstract}
Open-vocabulary detection (OVD) is a challenging task to detect and classify objects from an unrestricted set of categories, including those unseen during training. Existing open-vocabulary detectors are limited by complex visual-textual misalignment and long-tailed category imbalances, leading to suboptimal performance in challenging scenarios. To address these limitations, we introduce \textbf{MQADet}, a universal paradigm for enhancing existing open-vocabulary detectors by leveraging the cross-modal reasoning capabilities of multimodal large language models (MLLMs). MQADet functions as a plug-and-play solution that integrates seamlessly with pre-trained object detectors without substantial additional training costs. Specifically, we design a novel three-stage Multimodal Question Answering (MQA) pipeline to guide the MLLMs to precisely localize complex textual and visual targets while effectively enhancing the focus of existing object detectors on relevant objects. To validate our approach, we present a new benchmark for evaluating our paradigm on four challenging open-vocabulary datasets, employing three state-of-the-art object detectors as baselines. Experimental results demonstrate that our proposed paradigm significantly improves the performance of existing detectors, particularly in unseen complex categories, across diverse and challenging scenarios. To facilitate future research, we will publicly release our code.
\end{abstract}

\begin{figure}[htb!]
  \centering
  \begin{subfigure}[b]{0.48\linewidth}
      \centering
      \includegraphics[width=\linewidth]{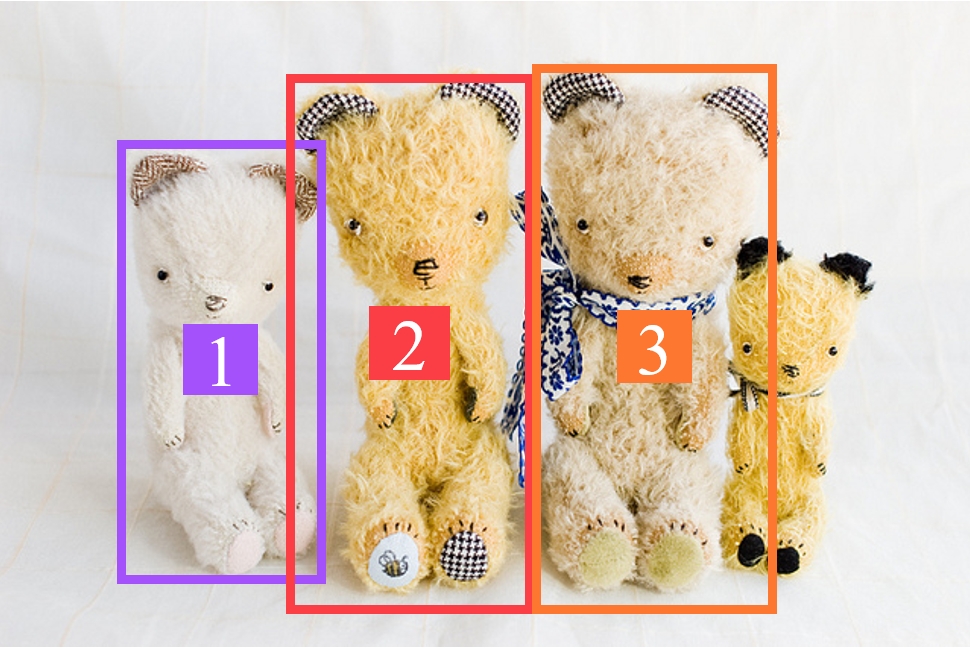}
      \caption{Grounding DINO}
      \label{subfig1}
  \end{subfigure}
  \begin{subfigure}[b]{0.48\linewidth}
      \centering
      \includegraphics[width=\linewidth]{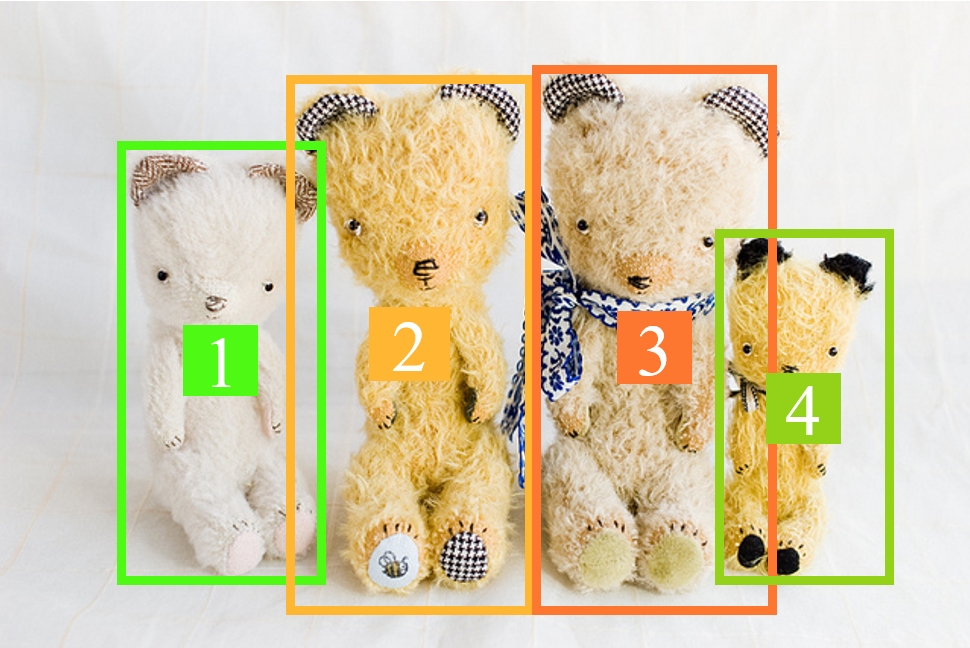}
      \caption{YOLO-World}
      \label{subfig2}
  \end{subfigure}
  \begin{subfigure}[b]{0.48\linewidth}
      \centering
      \includegraphics[width=\linewidth]{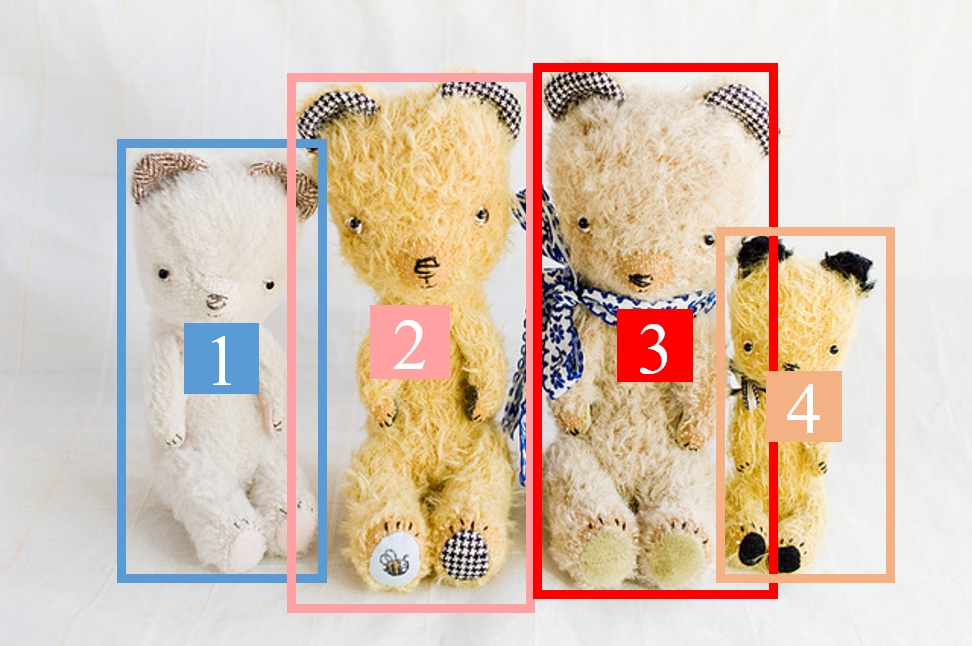}
      \caption{OmDet-Turbo}
      \label{subfig3}
  \end{subfigure}
  \begin{subfigure}[b]{0.48\linewidth}
      \centering
      \includegraphics[width=\linewidth]{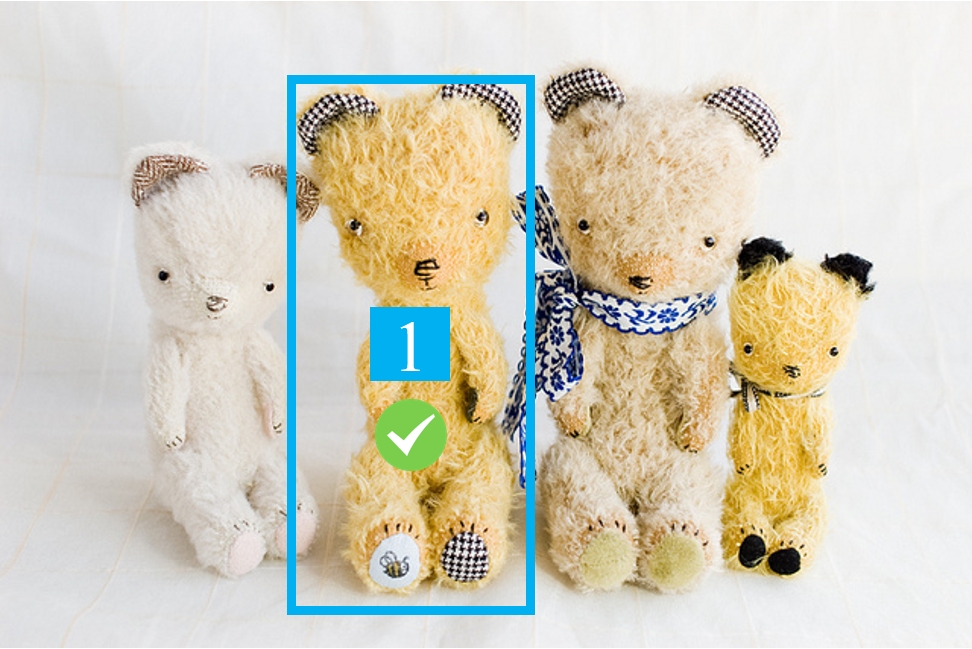}
      \caption{MQADet}
      \label{subfig4}
  \end{subfigure}
  \caption{Example of OVD in challenging scenarios. The detection target in this case is described as "a teddy bear with a checkered design on one foot and a bumble bee design on the other foot . the bear also has the checkered design over its ' ears". Comparison with previous OV detectors (e.g., Grounding DINO, YOLO-World, and OmDet-Turbo), MQADet significantly improves detection accuracy for objects described by complex textual queries.}
  \label{fig1}
\end{figure}

\section{Introduction}
Object detection is a fundamental task in the computer vision community, serving as a cornerstone for applications in image analysis, robotics, and autonomous driving~\cite{survey,objectrobotic,AD}. In recent years, significant advancements in deep learning architectures have enabled numerous studies~\cite{fasternn,diffusiondet,detrs} to achieve remarkable improvements in object detection accuracy. However, most existing detectors remain inherently constrained by a predefined set of object categories, such as the 80 classes defined in the COCO dataset~\cite{coco}. These detectors are limited to recognizing only the categories they were explicitly trained on, and expanding to new categories often requires extensive human-annotated data and carefully designed retraining strategies. Building on the proven effectiveness of MLLMs in multimodal reasoning tasks~\cite{li2024seed,cui2024survey}, recent research~\cite{liu2025grounding,cheng2024yolo,zhao2024real} has focused on extending these capabilities to open-vocabulary (OV) detection. While these approaches have improved detection accuracy in OV scenarios, they still face two significant challenges. First, their ability to align complex visual-textual information remains limited. As illustrated in Figure~\ref{fig1}, when tasked with detecting objects in multi-object scenarios described by complex textual queries: "a teddy bear with a checkered design on one foot and a bumble bee design on the other foot . the bear also has the checkered design over its ' ears", current state-of-the-art OVD methods, including Grounding DINO~\cite{liu2025grounding}, YOLO-World~\cite{cheng2024yolo}, and OmDet-Turbo~\cite{zhao2024real} fail to detect the queried teddy bear accurately. This shortcoming underscores their difficulty in processing intricate textual descriptions and achieving effective visual-textual alignment, ultimately limiting their reasoning capabilities across diverse object attributes. Second, these methods also demand substantial computational resources during training, which constrains their scalability and practical deployment.

To address the aforementioned challenges, this paper proposes a Plug-and-Play Paradigm, MQADet, which effectively enhances the detection accuracy of existing OV detectors for complex textual queries without requiring additional training resources. MQADet employs a three-stage Multimodal Question Answering pipeline: (1) Text-Aware Subject Extraction (TASE) leverages the advanced text understanding capabilities of MLLMs to extract key subjects and corresponding features from intricate textual descriptions; (2) Text-Guided Multimodal Object Positioning (TMOP) utilizes these extracted cues to guide existing OV detectors for precise object localization and detection; and (3) MLLMs-Driven Optimal Object Selection (MOOS) ensures fine-grained visual-textual alignment through MLLMs, optimizing object selection. This design effectively addresses the challenges of textual complexity and visual-textual alignment, significantly improving detection performance in OV scenarios. In summary, the contributions of this paper are threefold:
\begin{itemize}
    \item We propose a Plug-and-Play Paradigm, MQADet, which seamlessly integrates with pre-trained object detectors without incurring substantial additional training costs. MQADet introduces a three-stage Multimodal Question Answering (MQA) pipeline, effectively enhancing the detection accuracy of existing detectors in open-vocabulary scenarios with complex textual queries.
    \item We present a new benchmark designed to evaluate the performance of three existing OV detectors as well as our proposed MQADet paradigm when applied to these detectors across four challenging open-vocabulary datasets. Additionally, we provide a comprehensive analysis and comparison, offering valuable insights into MQADet's effectiveness across diverse textual queries and object categories.
    \item Experimental results demonstrate that MQADet effectively enhances open-vocabulary detection accuracy, achieving average improvements of 13\% on RefCOCO, 9\% on RefCOCO+, 20\% on RefCOCOg, and 27\% on Ref-L4. These results highlight the robustness and effectiveness of MQADet in addressing complex textual reasoning and alignment challenges.
\end{itemize}

\begin{figure*}[h!]
\centering
\includegraphics[width=0.8\textwidth]{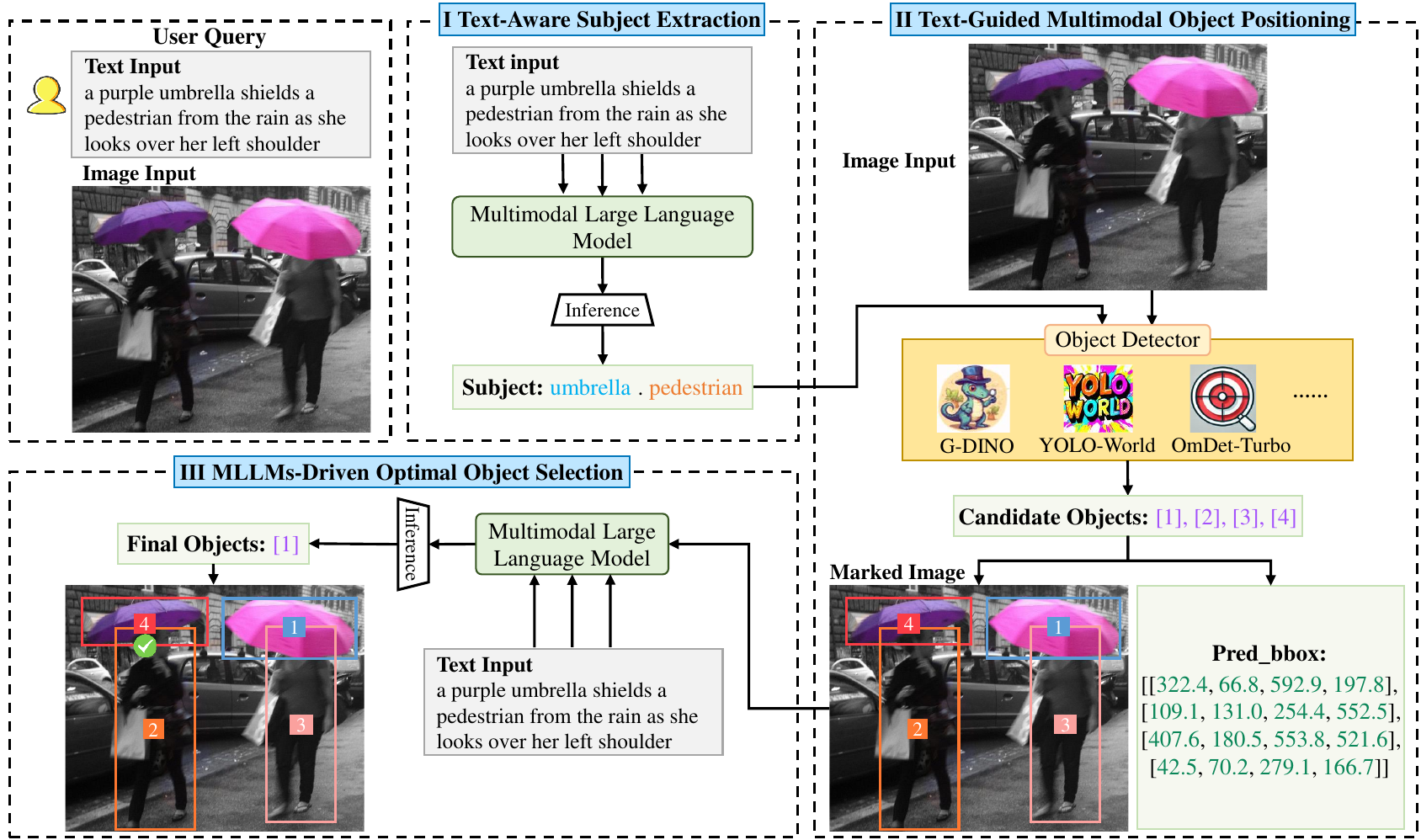}
\caption{An overview of the proposed MQADet paradigm, which consists of three Multimodal Question Answering (MQA) stages: Text-Aware Subject Extraction (TASE), Text-Guided Multimodal Object Positioning (TMOP), and MLLMs-Driven Optimal Object Selection (MOOS).}
\label{our}
\end{figure*}

\section{Related Work}
\subsection{Open-Vocabulary Detection}
Open-vocabulary detection (OVD) aims to generalize beyond the limited number of base classes labeled during the training phase and detect arbitrary classes. CLIP~\cite{pmlr-v139-radford21a} leverages cross-modal contrastive learning on large-scale image-text datasets to map text and images into the same embedding space, enabling zero-shot transfer to OV detection tasks. ViLD~\cite{zhong2022regionclip} is trained through visual and linguistic knowledge distillation, refining the knowledge obtained from CLIP inference into two-stage detectors to enhance zero-shot object detection. Region-CLIP~\cite{zhong2022regionclip} extend CLIP to enable it to learn region-level visual representations, enhancing CLIP's ability in open-set object detection tasks. Inspired by self-supervised learning methods, Grounding DINO~\cite{liu2025grounding} follow the design principles of tight modality fusion based on DINO~\cite{zhang2022dino} and large-scale grounded pre-training for zero-shot transfer. YOLO-World~\cite{cheng2024yolo} proposes a new Re-parameterizable Vision-Language Path Aggregation Network (RepVL-PAN) and a region-text contrastive loss to enhance the interaction between visual and linguistic information, achieving reduced computational requirements without sacrificing performance. These approaches aim to fuse multimodal information through single-stage methods with a limited number of parameters, seeking to form end-to-end vision-language understanding capabilities. However, they struggle to address the modality alignment problem between complex textual descriptions and vision, which limits their zero-shot capabilities and results in poor language generalization. These models tend to underperform on datasets they haven't been specifically fine-tuned on, especially when it comes to understanding long sentences.

\subsection{Modality Information Fusion}
Accurate Open-Vocabulary Detection relies on effective multimodal information fusion, which requires precise alignment between visual and linguistic modalities. CLIP~\cite{pmlr-v139-radford21a} matches entire images with textual descriptions, but it fails to capture the fine-grained alignment between image regions and their corresponding textual descriptions. MEDet~\cite{chen2022open} and VL-PLM~\cite{zhao2022exploiting} achieve alignment between potential object regions and entire textual descriptions (often simple words representing categories) by introducing RPNs or class-agnostic proposal generators. However, these methods do not fully address the understanding of complex longer sentences, which remains a challenge in achieving a more nuanced and detailed alignment of vision and language. CoOp~\cite{zhou2022learning} recognized that subtle changes in textual descriptions can have a significant impact on the performance of vision language pre-training models. It introduced CoOp for prompt representation learning and automated prompt engineering specifically for pre-trained vision language models. DetPro~\cite{du2022learning} incorporates CoOp into OV object detection tasks, enabling automatic learning of prompt representations in pre-trained vision language models based on positive and negative proposals. TaskCLIP~\cite{chen2024taskclip} adopts a two-stage design, combining general object detection with VLM-based object selection. It improves alignment between object images and their visual attributes by employing a transformer-based aligner to recalibrate the embeddings of both modalities. The above methods for modality information alignment involve complex training and high resource demands, hindering real-world deployment.

\section{Method}
\subsection{Problem Formulation}
In this work, MQADet aims to reason out optimal objects in complex textual query scenarios by combining MLLMs and existing OV detectors through a Multimodal Question Answering (MQA) mechanism. Given a user query text input comprising $n$ words, denoted as $T = {w_1, ..., w_n}$, the text consists of object categories, noun phrases, or object descriptions. The MLLMs are first tasked with identifying the object subjects in the text input, denoted as ${os_r}_{r=1}^{m}$, where $m$ represents the number of object subjects in the textual query. The identified object subjects, along with the original image $I$, are then provided as inputs to the OV detectors $Dets$, which generate bounding boxes for detection candidate objects ${Boxes_r}$ and the corresponding object marks ${Marks_r}$, forming the marked image $MI$. Finally, the optimal detection objects ${ob}$ are obtained by leveraging MLLMs to effectively align the text input $T$ with the marked image $MI$.

\subsection{Architecture of MQADet}
MQADet is a plug-and-play paradigm for enhancing OVD tasks without pre-training, which provides a new perspective for the community to address complex visual-textual modality alignment by utilizing the visual perception and reasoning capabilities of MLLMs. MQADet is comprised of three MQA stages: Text-Aware Subject Extraction (TASE) (Section~\ref{extraction}), Text-Guided Multimodal Object Positioning (TMOP) (Section~\ref{candidate}), and MLLMs-Driven Optimal Object Selection (MOOS) (Section~\ref{selection}), as shown in Figure~\ref{our}. 

Given an image $I$ and a complex detection text input $T$ as a user query, the TASE stage is responsible for identifying the target subjects described in the complex text query, along with their corresponding features. The TMOP stage then utilizes a state-of-the-art object detector to generate candidate target bounding boxes and marks corresponding to these identified subjects. Finally, the MOOS stage bridges the logic gap between perception and reasoning, arriving at the optimal detection result by leveraging the MQA mechanism. The subsequent sections provide a detailed explanation of the design for each phase.

\begin{figure*}[h!]
\centering
\includegraphics[width=1.0\textwidth]{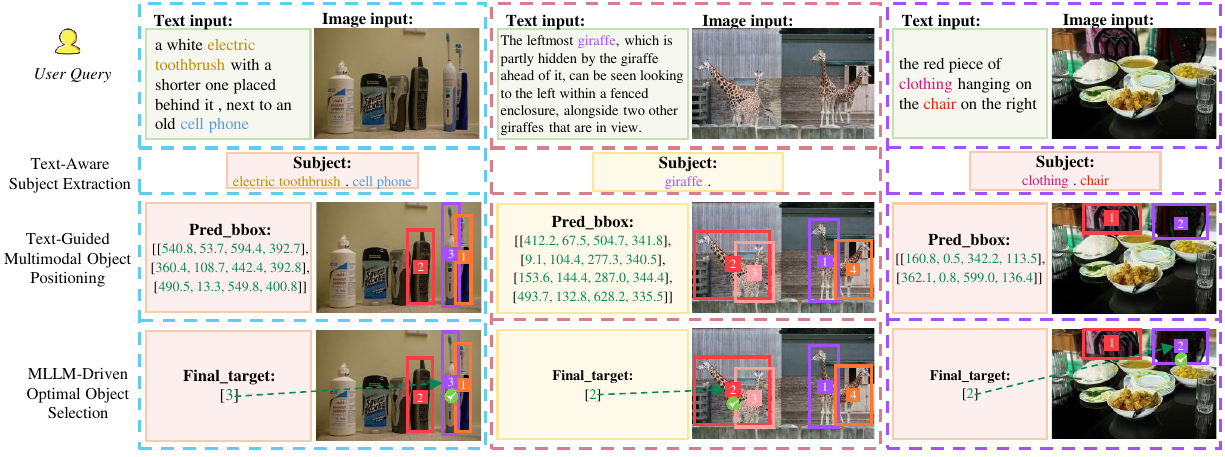}
\caption{Representative three cases using our MQADet paradigm. The specific inputs to MQADet and the corresponding output results across three phases are presented. Our paradigm effectively discerns a wider range of categories and reasons about the correct answers.}
\label{three-cases}
\end{figure*}

\subsection{Text-Aware Subject Extraction  (TASE)}\label{extraction}

OVD in real-world scenarios is inherently complex, requiring the orchestration of multiple subtasks to achieve its objective. For instance, users often aim to detect specific targets described with detailed sentences, such as “construction worker with a yellow helmet, reflective safety jacket, and pants,” rather than simple targets lacking descriptive features like “guy”, “car”, or “banana”. However, most existing OV detectors exhibit limited capability in addressing this challenge. Inspired by~\cite{sun2023multimodal}, we adopt a strategy of decomposing complex tasks into simpler, sequentially addressed subtasks. This decomposition is crucial for effectively tackling the challenges of textual complexity and forms the foundation of the proposed MQADet paradigm.

At the core of MQADet is the integration of Multimodal Large Language Models (MLLMs), which have demonstrated remarkable zero-shot and few-shot reasoning performance in Natural Language Processing tasks. To address the limitations of existing OV detectors, we propose Text-Aware Subject Extraction (TASE) as the initial stage of the MQADet. This stage leverages MLLMs and common-sense knowledge to parse and identify the target subjects, denoted as ${os_r}_{r=1}^{m}$, from the query text input $T$. The process is mathematically expressed as follows:
\begin{equation}
    {os_r}_{r=1}^{m} = MLLMs (T)
\label{subj}
\end{equation}

These subjects ${os_r}$, representing the target entities and their corresponding features from the user query, serve as the input for MQADet in the subsequent stage. This establishes a seamless flow between the subject extraction results and the following candidate positioning stage. The details of the prompts utilized for the MLLMs in this stage are provided in Appendix~\ref{Prompt_satge1}.

\subsection{Text-Guided Multimodal Object Positioning (TMOP)}\label{candidate}
In the previous stage, we obtained the target subjects and their corresponding features from the user query text. The TMOP stage processes these identified target subjects along with the input image $I$ to generate candidate-predicted boxes and assign corresponding object marks. 

Specifically, a state-of-the-art OV object detector (e.g., Grounding DINO, YOLO-World, or OmDet-Turbo) is integrated into this stage. This facilitates the automated identification of candidate bounding boxes $Boxes_r$ for each subject within the image, with numerical marks $Marks_r$ assigned to the center of each detection box being marked. The process is mathematically expressed as follows:
\begin{equation}
    \{Boxes_r, Marks_r\} = Dets (os_r, I)
\label{boxes_marks}
\end{equation}
\begin{equation}
    MI = I (Boxes, Marks)
\label{marked_img}
\end{equation}

where $Dets$ indicates detectors, $r$ corresponds to the $r$-th object subjects, $I$ denotes the original image of user query, $MI$ refers to the marked image. In particular, our paradigm allows on-demand integration of state-of-the-art target detectors without additional expensive training resources, a design that coincides with the current era of rapidly evolving large models. This approach highlights leveraging advanced detectors to efficiently focus on and capture a diverse range of potential targets in OV scenarios. The resulting marked images are then passed as input to the subsequent stage, where MLLMs-driven optimal object selection is performed.

\subsection{MLLMs-Driven Optimal Object Selection (MOOS)}\label{selection}
The MOOS stage ensures the fine-grained alignment between visual targets and complex longer language, achieving the optimal detection objects $ob$. Specifically, this object-selection stage is framed as a choice-based MQA task. The MLLMs are guided by detailed instruction prompts, effectively resolving challenging OV object detection problems by responding to a bounding box index. The process is mathematically formulated as follows:
\begin{equation}
    ob = MLLMs(T, MI)
\label{opt_objs2}
\end{equation}

This procedure endeavors to transfer MLLMs' vision-language reasoning capabilities to fulfill a more nuanced alignment of potential regions and descriptions in the given image, significantly boosting the detector's recognition capabilities in OV scenarios. In this work, GPT-4o and LLaVA were utilized as the MLLMs. Details of the prompts used for the MLLMs in this stage are provided in Appendix~\ref{Prompt_satge3}.

\section{Benchmark}

\subsection{Datasets and Evaluation Metrics}

\paragraph{Datasets.} To evaluate the detection performance of our paradigm in OV scenarios, we assess its zero-shot capabilities on four benchmark datasets: RefCOCO, RefCOCO+, RefCOCOg, and Ref-L4. These datasets, characterized by their intricate textual descriptions, are widely used in object detection.

\textbf{RefCOCO}~\cite{yu2016modeling}, \textbf{RefCOCO+}~\cite{yu2016modeling}, and \textbf{RefCOCOg}~\cite{nagaraja2016modeling} are designed to facilitate the understanding of natural language expressions referring to specific objects in images. RefCOCO+ specifically excludes locational prepositions, such as "on the right". In contrast, RefCOCOg includes spatial relations and features longer expressions. The average query lengths are 3.61, 3.53, and 8.43 words for RefCOCO, RefCOCO+, and RefCOCOg, respectively, reflecting the increasing complexity of their descriptions.

\textbf{Ref-L4}~\cite{chen2024revisiting} is a newly introduced benchmark dataset for object detection. It features a diverse range of 365 distinct object categories, with instance scales ranging from 30 to 3,767. The dataset is notable for its lengthy referring expressions, averaging 24.2 words, and features an extensive vocabulary of 22,813 unique words.

\paragraph{Evaluation Metrics.} We report Acc@0.25, Acc@0.5, and $\Delta$ to assess the object detection capabilities of different models, following previous works~\cite{peng2023kosmos,zhan2025griffon,chen2023shikra,zhang2023next}. Acc@0.25 and Acc@0.5 refer to accuracy metrics for bounding box predictions, where the Intersection-over-Union (IoU) with the ground-truth bounding boxes exceeds thresholds of 0.25 and 0.5, respectively, as expressed in Equation~\ref{IoU} and~\ref{acc}. The symbol $\Delta$ represents the improvement gains over baseline methods.
\begin{equation}
    IoU = \frac{area(B_p \cap B_{gt})}{area(B_p \cup B_{gt})}
\label{IoU}
\end{equation}

\begin{equation}
    Acc@IoU(T) = \frac{1}{N}\sum\limits_{i=1}^{N} 1(IoU_i \geq T)
\label{acc}
\end{equation}

where IoU is the ratio of the intersection area of the predicted bounding box $B_p$ and the ground truth bounding box $B_{gt}$ to the area of their union. $N$ denotes the total number of ground truth objects. The $1(IoU_i \geq T)$ is an indicator function, which is 1 if the IoU of the i-th prediction and ground truth exceeds or equals the threshold $T$, and 0 otherwise. $IoU_i$ denotes the IoU for the i-th ground truth and predicted bounding box. $T$ is the IoU threshold (0.25 or 0.5).

\subsection{Baselines}

As our baselines, we employ three state-of-the-art OV object detection models (Grounding DINO, YOLO-World, and OmDet-Turbo) and two MLLMs (GPT-4o and LLaVA-1.5).

\paragraph{Grounding DINO.} Grounding DINO~\cite{liu2025grounding} is an advanced open-set object detector that detects arbitrary objects based on human inputs, such as category names or referring expressions. It extends a closed-set object detection model by incorporating a text encoder, enabling open-set object detection with remarkable performance.

\paragraph{YOLO-World.} YOLO-World~\cite{cheng2024yolo} is a cutting-edge zero-shot object detection model. Unlike traditional YOLO detectors, it supports OV detection and accepts text input to recognize object instances in images without fine-tuning. The model follows the YOLO standard architecture while integrating a pre-trained CLIP text encoder, achieving both fast inference speeds and ease of deployment.

\paragraph{OmDet-Turbo.} OmDet-Turbo~\cite{zhao2024real} is a transformer-based OV detector designed for real-time applications. It effectively balances robust detection performance with high inference efficiency, excelling in OV scenarios and demonstrating strong zero-shot detection capabilities.

\paragraph{GPT-4o.} The advantages of GPT-4o\footnote{\url{https://openai.com/index/hello-gpt-4o/}} include: 1) real-time interaction with zero-latency dialogue and simultaneous processing of audio, visual, and text inputs; 2) speedy response, twice as fast as GPT-4 Turbo\footnote{\url{https://help.openai.com/en/articles/8555510-gpt-4-turbo}}, ideal for applications needing immediate responses; and 3) cross-modal capability, with superior audio, vision understanding, and reasoning, making it especially effective for visual grounding tasks.

\paragraph{LLaVA-1.5.} Large Language and Vision Assistant (LLaVA)~\cite{liu2024visual} is an end-to-end trained large multimodal model that connects a vision encoder and an LLM for general-purpose visual and language understanding. LLaVA-1.5~\cite{LLaVA1.5} makes simple modifications to LLaVA, namely, using CLIP-ViT-L-336px with an MLP projection and adding academic-task-oriented VQA data with response formatting prompts.

\begin{table*}[htb!]
\caption{Results comparison between MQADet and state-of-the-art detectors on RefCOCO/+/g, and Ref-L4. MLLM employs GPT-4o, while object detectors utilize Grounding DINO~\protect\cite{liu2025grounding}, YOLO-World~\protect\cite{cheng2024yolo}, and OmDet-Turbo~\protect\cite{zhao2024real}. The evaluation metrics are Acc@0.5, Acc@0.25, and $\Delta$. Numbers in {\color{red}red} point out the improvement gains between detector baselines and MQADet. }
    \label{table1}
    \setlength{\tabcolsep}{1pt}
    \centering
    \begin{tabular}{cccccccccccccccc}
    \toprule
\multirow{2}{*}{\textbf{Mehtod}}  & \multirow{2}{*}{\textbf{Metric}} & \multicolumn{4}{c}{\textbf{RefCOCO}}  & \multicolumn{4}{c}{\textbf{RefCOCO+}} & \multicolumn{3}{c}{\textbf{RefCOCOg}}  & \multicolumn{2}{c}{\textbf{Ref-L4}}\\ 
    \cmidrule(lr){3-6}\cmidrule(lr){7-10}\cmidrule(lr){11-13}\cmidrule(lr){14-15}   
    & & train & val & testA  & testB  & train  & val & testA & testB & train  & val  & test  & val  & test\\ 
    \midrule
    \midrule
\multirow{2}{*}{G-DINO}  
    &Acc@0.25 &  48.00&	48.95&	49.83&	40.50&	48.14&	49.66&	50.58&	43.51&	42.21&	40.76&	41.96&	17.40&	17.19\\ 
    \cmidrule{2-15} 
    & Acc@0.5 &  43.14&	42.85&	45.07&	36.69&	41.77&	41.56&	43.98&	37.51&	39.43&	38.18&	39.24&	16.66&	16.34\\ 
    \midrule
\multirow{4}{*}{Ours (G-DINO)}   
    &Acc@0.25  &64.70&	66.59&	64.01&	67.20&  57.35& 57.29& 55.07& 56.87&	66.52&	66.10&	67.91 &63.71& 64.21\\ 
    & $\Delta$ & {\color{red}+16.7} &  {\color{red}+17.64} & {\color{red}+14.18} &  {\color{red}+26.7}& {\color{red}+9.21}& {\color{red}+7.63} & {\color{red}+4.49} & {\color{red}+13.36} & {\color{red}+24.31}& {\color{red}+25.34}& {\color{red}+25.95}& {\color{red}+46.31}&  {\color{red}+47.02}\\
    \cmidrule{2-15} 
    & Acc@0.5  & 58.92 &60.47& 60.03& 61.70& 50.62&49.50& 48.51&	50.18&	61.58&	61.45&	62.90& 59.35& 59.34\\
    & $\Delta$ & {\color{red}+15.78} &  {\color{red}+17.62} & {\color{red}+14.96} & {\color{red}+25.01} & {\color{red}+8.85} & {\color{red}+7.94} & {\color{red}+4.53} & {\color{red}+12.67} & {\color{red}+22.15} & {\color{red}+23.27} & {\color{red}+23.66} & {\color{red}+42.69} & {\color{red}+43.0}\\
    \midrule
    \midrule
\multirow{2}{*}{YOLO-World} & 
    Acc@0.25& 38.79& 38.15&	42.70&	32.97&	39.24&	37.82&	38.20&	35.32&	42.44&	40.11&	43.05&	28.76&	29.94\\ 
    \cmidrule{2-15} 
    & Acc@0.5  &34.09&	32.65&	38.36&	28.47&	33.56&	31.06&	33.77&	30.65&	38.43&	36.99&	38.51&	25.25&	26.56\\ 
    \midrule
\multirow{4}{*}{Ours (YOLO-World)}  & 
    Acc@0.25  & 63.72 &62.79& 60.59&	62.13 & 56.15& 56.97& 52.91& 55.47& 66.15& 62.50 &65.57 & 62.98& 57.75\\
    & $\Delta$ & {\color{red}+24.93} &  {\color{red}+24.64} & {\color{red}+17.89} &{\color{red}+29.16} &{\color{red}+16.91} &{\color{red}+19.15} &{\color{red}+14.71} &{\color{red}+20.15} &{\color{red}+23.71} &{\color{red}+22.39} &{\color{red}+22.52} &{\color{red}+34.22} &{\color{red}+27.81} &\\
    \cmidrule{2-15} 
    & Acc@0.5 & 57.98 &56.81&55.28&	55.65  &49.76 &48.31&	46.88 &48.84 &61.17	&57.55&	60.44& 57.86& 53.22  \\ 
    & $\Delta$ & {\color{red}+23.89} &  {\color{red}+24.16} & {\color{red}+16.92} & {\color{red}+27.18} & {\color{red}+16.2} & {\color{red}+17.25} & {\color{red}+13.11} & {\color{red}+18.19} & {\color{red}+22.74} & {\color{red}+20.56} & {\color{red}+21.93} & {\color{red}+32.61} & {\color{red}+26.66}\\
    \midrule
    \midrule
\multirow{2}{*}{OmDet-Turbo} & 
    Acc@0.25& 49.62& 48.87&	55.44&	41.38&	48.07&	46.84&	49.03&	44.09&	46.02&	42.81&	45.01&	32.29&	32.16\\ 
    \cmidrule{2-15} 
    & Acc@0.5&  46.53&	45.43&	52.76&	37.06&	44.57&	42.96&	46.03&	37.94&	40.79&	38.27&	39.07&	28.67&	28.95\\ 
    \midrule
\multirow{4}{*}{Ours (OmDet-Turbo)}  & 
    Acc@0.25 & 62.04&58.34	&64.48&	50.56&54.59& 54.34&	55.91& 51.59&59.20&56.82& 57.62& 56.94 & 54.06\\ 
    & $\Delta$ & {\color{red}+12.42} & {\color{red}+9.47} & {\color{red}+9.04} &
    {\color{red}+9.18} & {\color{red}+6.52} &
    {\color{red}+7.5} & {\color{red}+6.88} &
    {\color{red}+7.5} & {\color{red}+13.18} &
    {\color{red}+14.01} & {\color{red}+12.61} &
    {\color{red}+24.65} & {\color{red}+21.9} &\\
    \cmidrule{2-15} 
    & Acc@0.5 & 58.07&53.77&61.39&45.84&50.07&49.46	&53.21&	46.47&54.55& 52.22&	52.90&	51.66&	49.55\\
    & $\Delta$ & {\color{red}+11.54} &  {\color{red}+8.34} & {\color{red}+8.63} & {\color{red}+8.78} & {\color{red}+5.5} & {\color{red}+6.5} & {\color{red}+7.18} & {\color{red}+8.53} & {\color{red}+13.76} & {\color{red}+13.95} & {\color{red}+13.83} & {\color{red}+22.99} & {\color{red}+20.6}\\
    \bottomrule
    \end{tabular}
\end{table*}

\section{Experiments}

\subsection{Implementation Details}
In our experiments, we separately adopt \textit{gpt-4o} and \textit{llava-v1.5-7b} as the reasoning MLLMs. In the Text-Guided Multimodal Object Positioning (TMOP) stage, we utilize three OV object detection models: Grounding DINO, YOLO-World, and OmDet-Turbo. For Grounding DINO, we set the box\_threshold and text\_threshold to 0.25, with the inference model configured to \textit{GroundingDINO-T}. For YOLO-World, the predicted model is \textit{YOLO-Worldv2-XL}. For OmDet-Turbo, we use \textit{OmDet-Turbo\_tiny\_SWIN\_T} as the inference model. In our experiments, specific implementation checkpoints are detailed in Appendix~\ref{model_details}.

For four benchmark datasets: RefCOCO, RefCOCO+, RefCOCOg, and Ref-L4, we implement uniform sampling, establishing a sampling ratio of 0.1. After sampling, the RefCOCO dataset includes 12,062 expressions in the train set, 565 in testA, 509 in testB, and 1,083 in the validation set. RefCOCO+ comprises 12,019 expressions in the train set, 572 in testA, 488 in testB, and 1,075 in the validation set. RefCOCOg contains 8,051 expressions in the train set, 960 in the test set, and 489 in the validation set. Ref-L4 consists of 3,192 expressions in the test set and 1,342 in the validation set. We perform the experiments using a single RTX 4090 GPU. The input and output results for the different phases of MQADet are detailed in Appendix~\ref{MQADet_details}.

\begin{table*}[htb!]
\caption{Ablation experiments on RefCOCO testA and Ref-L4 val datasets. MLLM employs GPT-4o, while object detectors utilize Grounding DINO~\protect\cite{liu2025grounding}, YOLO-World~\protect\cite{cheng2024yolo}, and OmDet-Turbo~\protect\cite{zhao2024real} in TMOP stage.}
\label{ablation}
    \setlength{\tabcolsep}{2pt}
    \centering
    \begin{tabular}{ccccccccc}
    \toprule
\multirow{2}{*}{\textbf{TASE Stage}} & \multicolumn{3}{c}{\textbf{TMOP Satge}}  & \multirow{2}{*}{\textbf{MOOS Satge}} & \multicolumn{2}{c}{\textbf{\makecell{RefCOCO\\testA}}} &  \multicolumn{2}{c}{\textbf{\makecell{Ref-L4\\val}}}\\ 
    \cmidrule(lr){2-4}\cmidrule(lr){6-7}\cmidrule(lr){8-9}   
    & G-DINO  & YOLO-World & OmDet-Turbo & & Acc@0.25 &Acc@0.5& Acc@0.25&Acc@0.5\\ 
\midrule
    & \checkmark &  &  &\checkmark & 44.68&	40.78&	54.97&	44.76\\ 
    &  &  \checkmark & &\checkmark & 54.69&	48.85&	36.74&	43.04\\ 
    & & & \checkmark &  \checkmark & 56.31&	54.90&	49.96&	43.07\\ 
\midrule
    \checkmark & \checkmark&  & & &43.80& 40.42&	37.82&	36.28\\ 
    \checkmark  &  & \checkmark &  & &43.79&	39.67&	40.15&	37.65\\ 
    \checkmark  & &  & \checkmark & &50.15&	47.00&	44.52&	40.87\\ 
\midrule
    \checkmark & \checkmark &  & & \checkmark & \textbf{64.01} & \textbf{60.03} & \textbf{63.71} & \textbf{59.35} \\ 
    \checkmark  & &\checkmark & &  \checkmark & \textbf{60.59}&	\textbf{55.28} & \textbf{62.98} &	\textbf{57.86}\\ 
    \checkmark  & & &\checkmark &  \checkmark &\textbf{64.48}& \textbf{61.39}& \textbf{56.94}&	\textbf{51.66}\\ 
    \toprule
    \end{tabular}
\end{table*}

\begin{figure*}[htb!]
\centering
\includegraphics[width=1.0\textwidth]{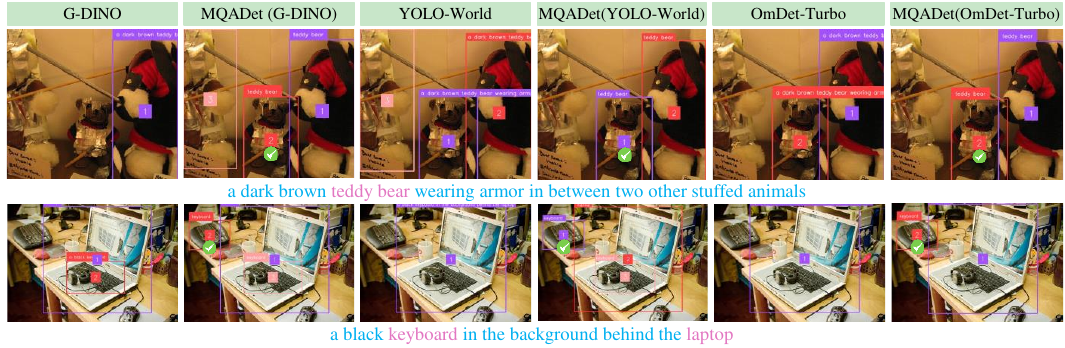}
\caption{The visualization results of Grounding DINO, YOLO-World, OmDet-Turbo, and MQADet, with GPT-4o employed as the MLLM. \textcolor{mypink}{Pink} words indicate the subjects identified from the user query. Please zoom in to view the detailed labels.}
\label{visual}
\end{figure*}

\begin{figure}[htb!]
  \centering
  \begin{subfigure}[b]{0.49\linewidth}
      \centering
      \includegraphics[width=\linewidth]{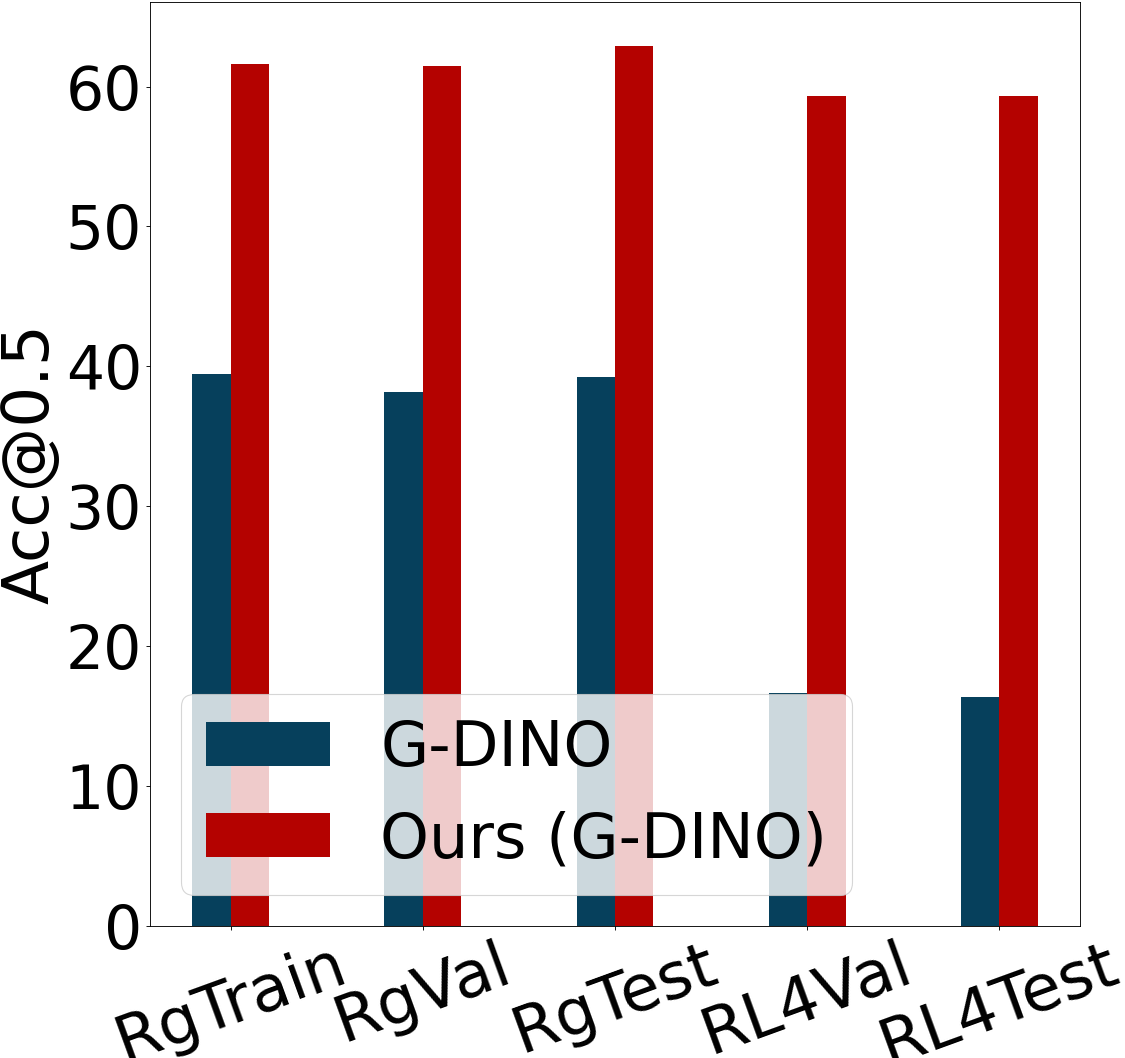}
      \caption{Grounding DINO}
      \label{subfig1}
  \end{subfigure}
  \begin{subfigure}[b]{0.49\linewidth}
      \centering
      \includegraphics[width=\linewidth]{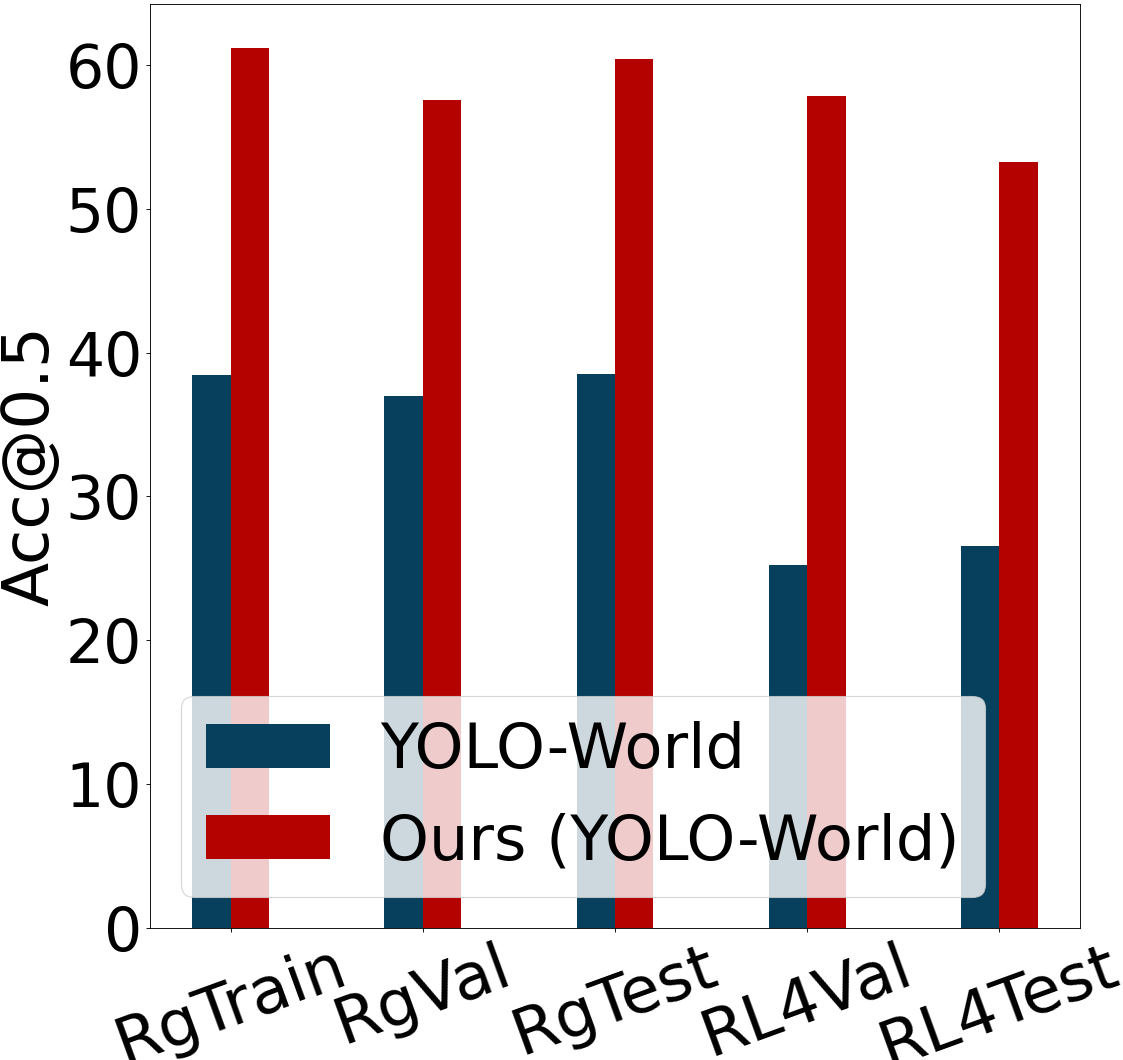}
      \caption{YOLO-World}
      \label{subfig2}
  \end{subfigure}
\caption{Performance comparison of MQADet and detectors on challenging RefCOCOg and Ref-L4. MLLM employs GPT-4o, while object detectors utilize Grounding DINO and YOLO-World. The evaluation metric is Acc@0.5. RgTrain, RgVal, RgTest, RL4Val, RL4Test refer to the RefCOCOg train, RefCOCOg val, RefCOCOg test, Ref-L4 val, and Ref-L4 test datasets, respectively.}
\label{curve}
\end{figure}

\subsection{Main Results}

\paragraph{Performance on GPT-4o.} Comparisons between MQADet employing GPT-4o as the MLLM and three advanced detectors on RefCOCO, RefCOCO+, RefCOCOg, and Ref-L4 are presented in Table~\ref{table1}. The results demonstrate that MQADet achieves superior performance in the OVD task across all four benchmarks under the same settings. Specifically, MQADet delivers significant improvements on RefCOCO, RefCOCO+, and RefCOCOg benchmarks and also excels on the challenging Ref-L4 dataset with longer descriptions. Additionally, to evaluate the generalizability of our paradigm, we conducted further assessments on three advanced detectors. The proposed paradigm consistently and significantly enhances the OV detection performance of these detectors across various benchmarks. As shown in Table~\ref{table1}, MQADet achieves remarkable improvements on the Ref-L4 test dataset. At Acc@0.5, Grounding DINO improves by 43.0\%, YOLO-World by 26.66\%, and OmDet-Turbo by 20.6\%. Similarly, at Acc@0.25, Grounding DINO achieves a 47.02\% increase, YOLO-World 27.81\%, and OmDet-Turbo 21.9\%. These results highlight the paradigm's generalizability and adaptability across different OV detection models. Moreover, the findings indicate that, compared with these detectors, MQADet can more effectively tackle the challenges posed by textual complexity and fine-grained misalignment of visual regions and complex descriptions by leveraging the vision-language reasoning capabilities of MLLM.

Figure~\ref{curve} illustrates the performance comparison of MQADet and detectors on challenging RefCOCOg and Ref-L4. MLLM employs GPT-4o, and object detectors utilize Grounding DINO and YOLO-World. MQADet consistently achieves significant performance gains on both benchmarks.

\paragraph{Performance on LLaVA-1.5.} To further validate the OVD capabilities and transferability of MQADet, we conducted additional evaluations using LLaVA-1.5 as the MLLM. The comparisons and analysis between MQADet and the advanced detectors on RefCOCO, RefCOCO+, RefCOCOg, and Ref-L4 are presented in Appendix~\ref{llava_result}.

\subsection{Ablation Experiments}

In ablation experiments, we remove certain stages in different variants of our MQADet paradigm, a three-stage MQA pipeline, to verify its effectiveness. Ablation experiments on RefCOCO testA and Ref-L4 val datasets are shown in Table~\ref{ablation}. MLLM employs GPT-4o, and object detectors utilize Grounding DINO, YOLO-World, and OmDet-Turbo in stage II. Notably, MQADet experiences a remarkable improvement and excels with the highest score on RefCOCO testA and Ref-L4 val datasets, surpassing the performance of the other variants. In contrast, removing stage I or stage III significantly degrades the OV detection results, highlighting the crucial role of parsing and identifying subjects from complex textual queries, as well as achieving nuanced alignment between intricate descriptions and visual targets, utilizing MLLMs. This result demonstrates the performance advantage and effectiveness of MQADet in diverse OV applications.

\subsection{Visualizations}

Figure~\ref{visual} shows the visualization results of Grounding DINO, YOLO-World, OmDet-Turbo, and MQADet, with GPT-4o employed as the MLLM. (1) Row 1 user query: woman in brown shirt with white purse hugging bear. (2) Row 2 user query: a black keyboard in the background behind the laptop. The visualization results demonstrate that MQADet enables the detectors to focus on more categories by identified subject cues and excels in detecting objects with complex descriptions by leveraging the vision-language reasoning capabilities of MLLMs.

\section{Conclusion}

In this paper, we review that existing open-vocabulary detectors are constrained by textual complexity and fine-grained misalignment of visual regions and complex descriptions, and propose a novel three-stage MQA paradigm termed MQADet. The MQADet contains three stages: Text-Aware Subject Extraction (TASE), Text-Guided Multimodal Object Positioning (TMOP), and MLLMs-Driven Optimal Object Selection (MOOS), to effectively and significantly boost the detection performance of existing OV detectors for complex textual queries. Extensive experiments prove that our paradigm outperforms existing state-of-the-art detectors across four benchmarks.

\bibliographystyle{named}
\bibliography{ijcai25}

\end{document}


\maketitle

\section{Overview}
In this appendix, more model details, main experiment results, MQADet details, and qualitative results are provided, which are organized as follows:
\begin{itemize}
    \item Sec.~\ref{model_details} provides the specific models and corresponding checkpoints used in our experiments.
    \item Sec.~\ref{llava_result} presents additional main experiment results and analysis using LLaVA-1.5 as the MLLM.
    \item Sec.~\ref{MQADet_details} details the inputs and outputs of MQADet, along with the corresponding prompts for each phase, using a concrete example.
    \item Sec.~\ref{qualitative_results} offers additional qualitative results for GPT-4o and LLaVA-1.5, where our method achieves superior performance.
\end{itemize}

\section{Model Details}\label{model_details}

The specific models and corresponding checkpoints employed in the MQADet paradigm are presented in
Table~\ref{checkpoints}. In our experiments, all the open-source models are loaded directly from HuggingFace or GitHub repository. The detailed trained weights are selected as follows:

\begin{table}[h!]
\newcommand{\tabincell}[2]{\begin{tabular}{@{}#1@{}}#2\end{tabular}}
\caption{Checkpoints for the different models.}
\label{checkpoints}
\setlength{\tabcolsep}{2pt}
\centering
    \begin{tabular}{cc}
    \toprule
    \textbf{Model} &  \textbf{Checkpoints}\\  
    \midrule
    Grounding DINO & groundingdino\_swint\_ogc.pth\\
    YOLO-World & \tabincell{c}{yolo\_world\_v2\_xl\_obj365v1\_goldg\\\_cc3mlite\_pretrain.pth}\\
    OmDet-Turbo & OmDet-Turbo\_tiny\_SWIN\_T.pth\\
    LLaVA-1.5 & liuhaotian/llava-v1.5-7b\\
    \bottomrule
    \end{tabular}
\end{table}

GPT-4o is not open-sourced, consequently, the models are not identifiable. The GPT-4o API released by OpenAI is utilized to evaluate all benchmark datasets.

\section{More Main Results}\label{llava_result}

\begin{table*}[htb!]
\caption{Results comparison between MQADet and state-of-the-art dectors on RefCOCO/+/g, and Ref-L4. MLLM employs LLaVA-1.5~\protect\cite{liu2024improved}, while object detectors utilize Grounding DINO~\protect\cite{liu2025grounding}, YOLO-World~\protect\cite{cheng2024yolo}, and OmDet-Turbo~\protect\cite{zhao2024real}. The evaluation metrics are Acc@0.5, Acc@0.25, and $\Delta$. Numbers in {\color{red}red} point out the improvement gains between detectors and MQADet.}
    \label{table2}
    \setlength{\tabcolsep}{1pt}
    \centering
    \begin{tabular}{cccccccccccccccc}
    \toprule
\multirow{2}{*}{\textbf{Mehtod}}  & \multirow{2}{*}{\textbf{Metric}} & \multicolumn{4}{c}{\textbf{RefCOCO}}  & \multicolumn{4}{c}{\textbf{RefCOCO+}} & \multicolumn{3}{c}{\textbf{RefCOCOg}}  & \multicolumn{2}{c}{\textbf{Ref-L4}}\\ 
    \cmidrule(lr){3-6}\cmidrule(lr){7-10}\cmidrule(lr){11-13}\cmidrule(lr){14-15}   
    & & train & val & testA  & testB  & train  & val & testA & testB & train  & val  & test  & val  & test\\ 
    \midrule
    \midrule
\multirow{2}{*}{G-DINO}  
    &Acc@0.25 &  48.00&	48.95&	49.83&	40.50&	48.14&	49.66&	50.58&	43.51&	42.21&	40.76&	41.96&	17.40&	17.19\\ 
    \cmidrule{2-15} 
    & Acc@0.5 &  43.14&	42.85&	45.07&	36.69&	41.77&	41.56&	43.98&	37.51&	39.43&	38.18&	39.24&	16.66&	16.34\\ 
    \midrule
\multirow{4}{*}{Ours (G-DINO)}   
    &Acc@0.25&58.75&  53.92& 56.46& 52.26 & 55.16& 53.58 & 56.12& 51.64& 67.54& 66.05& 67.40&	54.40& 45.21\\ 
    & $\Delta$ & {\color{red}+10.75} &  {\color{red}+4.97}  & {\color{red}+6.63} &  {\color{red}+11.76} &{\color{red}+7.02} &  {\color{red}+3.92} & {\color{red}+5.54} &  {\color{red}+8.13} &{\color{red}+25.33} &  {\color{red}+25.29} &{\color{red}+25.44} &  {\color{red}+37.0}&  {\color{red}+28.02}\\
    \cmidrule{2-15} 
    & Acc@0.5 & 51.79 & 46.45& 51.33& 45.19& 47.16&	44.09 &	49.13&	43.03& 61.14 & 59.51&	61.15&	49.85 & 41.13 \\
    & $\Delta$ & {\color{red}+8.65} &  {\color{red}+3.6} & {\color{red}+6.26} &  {\color{red}+8.5} &{\color{red}+5.39} &  {\color{red}+2.53} &{\color{red}+5.15} &  {\color{red}+5.52} &{\color{red}+21.71} &  {\color{red}+21.33} &{\color{red}+21.91} &  {\color{red}+33.19} &{\color{red}+24.79} \\
    \midrule
    \midrule
\multirow{2}{*}{YOLO-World} & 
    Acc@0.25& 38.79& 38.15&	42.70&	32.97&	39.24&	37.82&	38.20&	35.32&	42.44&	40.11&	43.05&	28.76&	29.94\\ 
    \cmidrule{2-15} 
    & Acc@0.5  &34.09&	32.65&	38.36&	28.47&	33.56&	31.06&	33.77&	30.65&	38.43&	36.99&	38.51&	25.25&	26.56\\ 
    \midrule
\multirow{4}{*}{Ours (YOLO-World)} & 
    Acc@0.25  & 55.66&	46.63&	45.84&	43.22&	54.12&	51.72&	50.52&	45.29&	63.01&	50.31&	64.38&	53.13&	34.77\\
    & $\Delta$ & {\color{red}+16.87} &  {\color{red}+8.48} & {\color{red}+3.14} &
    {\color{red}+10.25} & {\color{red}+14.88} &
    {\color{red}+13.9} & {\color{red}+12.32} &
    {\color{red}+9.97} & {\color{red}+20.57} &{\color{red}+10.2} & {\color{red}+21.33} &
    {\color{red}+24.37} & {\color{red}+4.83} \\
    \cmidrule{2-15} 
    & Acc@0.5 &  48.97&	39.34&	40.71&	36.74&	45.87&	42.60&	44.76&	36.07&	56.49&	44.99&	56.88&	48.14&	31.42\\ 
    & $\Delta$ & {\color{red}+14.88} &  {\color{red}+6.69} & {\color{red}+2.35} &
    {\color{red}+8.27} & {\color{red}+12.31} &
    {\color{red}+11.54} & {\color{red}+10.99} &
    {\color{red}+5.42} & {\color{red}+18.06} &
    {\color{red}+8.0} & {\color{red}+18.37} &
    {\color{red}+22.89} & {\color{red}+4.86}\\
    \midrule
    \midrule
\multirow{2}{*}{OmDet-Turbo} & 
    Acc@0.25& 49.62& 48.87&	55.44&	41.38&	48.07&	46.84&	49.03&	44.09&	46.02&	42.81&	45.01&	32.29&	32.16\\ 
    \cmidrule{2-15} 
    & Acc@0.5&  46.53&	45.43&	52.76&	37.06&	44.57&	42.96&	46.03&	37.94&	40.79&	38.27&	39.07&	28.67&	28.95\\ 
    \midrule
\multirow{4}{*}{Ours (OmDet-Turbo)}  & 
    Acc@0.25 & 59.89&	58.17&	62.83&	50.88&	56.04&	54.05&	54.20&	47.95&	63.05&	66.67&	71.88&	52.83&	48.68\\ 
    & $\Delta$ & {\color{red}+10.27} &  {\color{red}+9.3} & {\color{red}+7.39} &{\color{red}+9.5} &{\color{red}+7.97} &{\color{red}+7.21} &{\color{red}+5.17} &{\color{red}+3.86} &{\color{red}+17.03} &{\color{red}+23.86} &{\color{red}+26.87} &
    {\color{red}+20.54} &{\color{red}+16.52} \\
    \cmidrule{2-15} 
    & Acc@0.5 & 53.65&	50.78&	57.52&	43.03&	49.54&	47.81&	50.87&	41.60&	56.35&	60.12&	62.60&	47.47&	42.95\\
    & $\Delta$ & {\color{red}+7.12} &  {\color{red}+5.35} & {\color{red}+4.76} &
    {\color{red}+5.97} & {\color{red}+4.97} &
    {\color{red}+4.85} & {\color{red}+4.84} &
    {\color{red}+3.66} & {\color{red}+15.56} &
    {\color{red}+21.85} & {\color{red}+23.53} &
    {\color{red}+18.8} & {\color{red}+14.0} \\
    \bottomrule
    \end{tabular}
\end{table*}

\paragraph{Performance on LLaVA-1.5.} To further validate the OVD capabilities and transferability of MQADet, we conducted additional evaluations using LLaVA-1.5 as the MLLM. Table~\ref{table2} compares MQADet, employing LLaVA as the MLLM, with state-of-the-art detectors on the RefCOCO, RefCOCO+, RefCOCOg, and Ref-L4 datasets. MQADet achieves the highest scores for both Acc@0.25 and Acc@0.5 across four benchmarks, significantly surpassing the performances of Grounding-DINO, YOLO-World, and OmDet-Turbo. The $\Delta$ results demonstrate substantial improvements across all datasets and detectors, underscoring the paradigm's effectiveness and generalizability. Notably, regardless of whether GPT or LLaVA is adopted as the MLLMs in the TASE stage and MOOS stage, MQADet’s three-stage MQA mechanism enables detectors to utilize extracted subject cues, dealing with intricate textual descriptions, and focus on a broader range of unknown categories, thereby enhancing OV detection. Additionally, neither the MLLMs (GPT-4o and LLaVA) in the TASE and MOOS stage nor the detectors in the TMOP stage require training of costly, large models. Instead, they operate as plug-and-play components, making MQADet a highly effective solution for diverse real-world detection scenarios.

\section{MQADet Details}\label{MQADet_details}

To ensure the consistency and accuracy of the experimental results, the same prompts were used for the MLLMs, with GPT-4o codenamed as \textit{gpt4o} and LLaVA as \textit{llava-v1.5-7b}, in both TASE stage and MOOS stage of MQADet. The following subsections describe in detail the specific inputs and outputs of MQADet, as well as the corresponding prompts for each phase, using a concrete example.

\begin{figure}[htb!]
\centering
\includegraphics[width=0.45\textwidth]{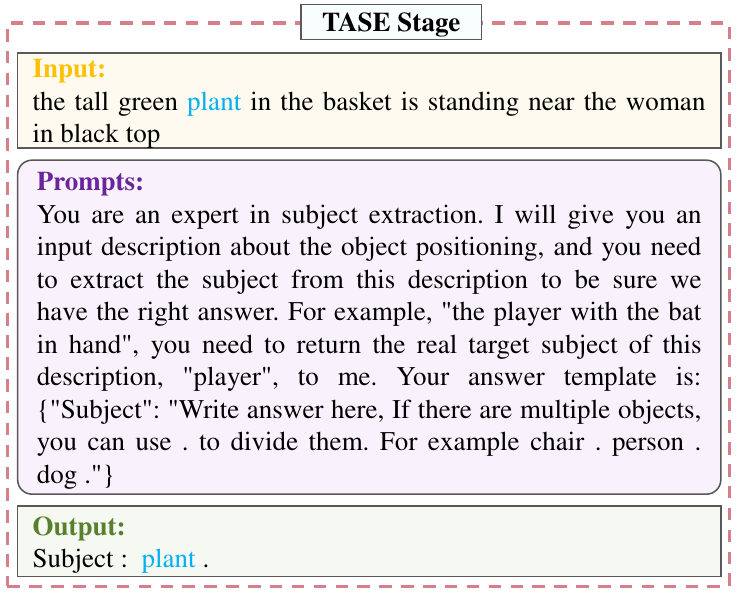}
\caption{Specific inputs, outputs, and corresponding prompts for the TASE stage.}
\label{s1}
\end{figure}  

\subsection{TASE Stage Details}\label{Prompt_satge1}

The input to the Text-Aware Subject Extraction (TASE) stage is a textual query with complex descriptions.
\begin{itemize}
    \item \textbf{Input}: The original text input of the user query. For example, \textit{"text input": "the tall green plant in the basket is standing near the woman in black top"}
\end{itemize}

In this stage, we use prompts, presented in Figure~\ref{s1}, to guide MLLMs (GPT-4o and LLaVA-1.5) in parsing and reasoning about the target subjects.
\begin{itemize}
    \item \textbf{Output}: Reasoned target subjects. For example, \textit{"subject": "plant ."}.
\end{itemize}



\subsection{TMOP Stage Details}\label{Prompt_satge3}

\begin{figure}[htb!]
\centering
\includegraphics[width=0.45\textwidth]{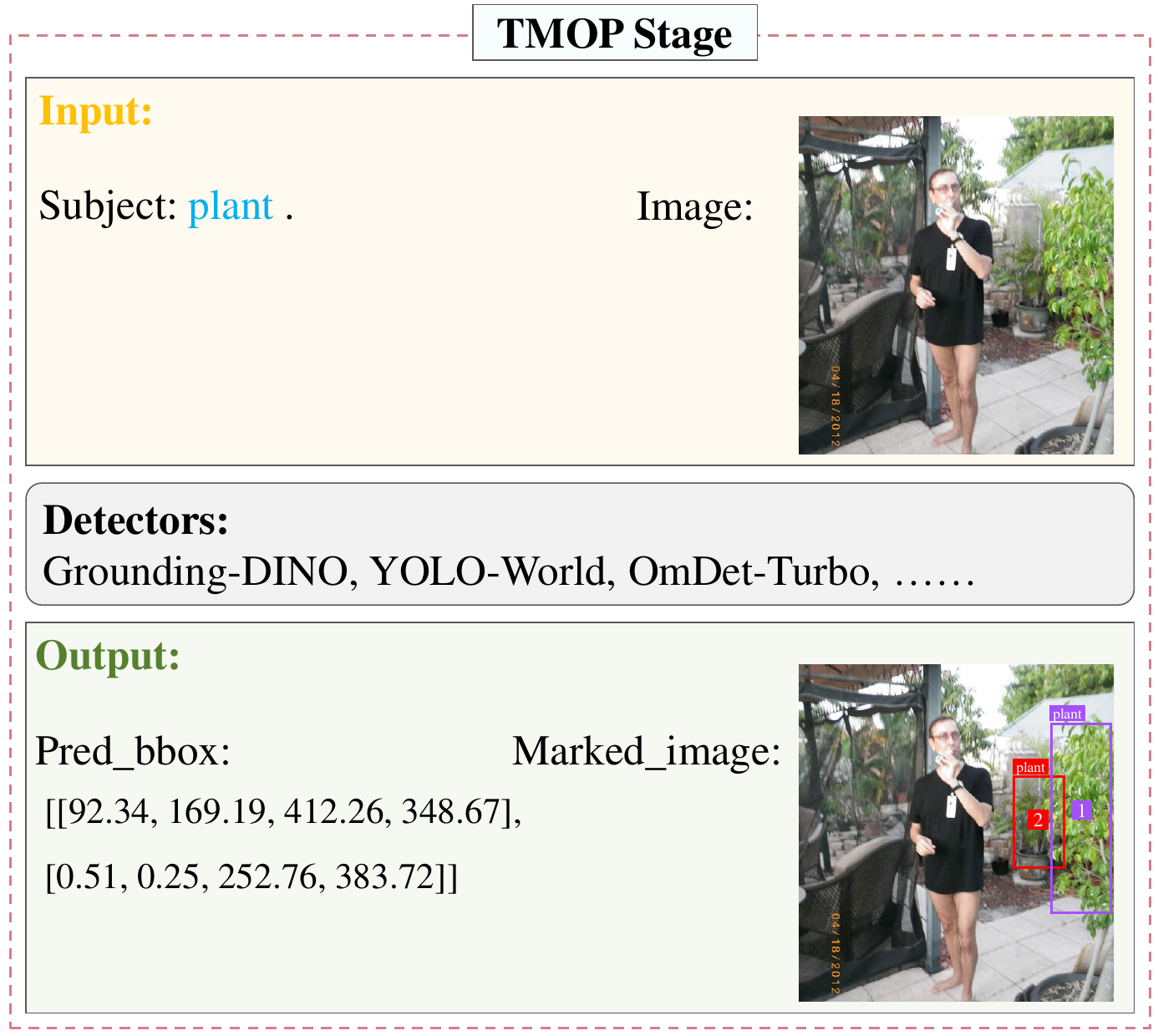}
\caption{Specific inputs, outputs, and detectors for the TMOP stage.}
\label{s2}
\end{figure}  

The outputs of the Text-Guided Multimodal Object Positioning (TMOP) stage are the total number of candidate bounding box coordinates, the corresponding number marks that OV object detectors can detect, and the marked image obtained, as shown in Figure~\ref{s2}.

\begin{itemize}
    \item \textbf{Input}: The subject prompts and the original input image are provided. For example, the subject prompts could be: \textit{"subject": "plant."}.
    \item \textbf{Output}: The candidate bounding box coordinates ([x\_min, y\_min, x\_max, y\_max]), the corresponding number of marks, and the resulting marked image. For example: 
    \begin{itemize}  
        \item \textit{"pred\_{bbox}": [[92.34, 169.19, 412.26, 348.67], [0.51, 0.25, 252.76, 383.72]]}
        \item \textit{marked image}
    \end{itemize}
\end{itemize}



\subsection{Prompts for MOOS Stage}\label{Prompt_satge3}

We take the original text input and the marked image as MLLMs-Driven Optimal Object Selection (MOOS) stage input.
\begin{itemize}
    \item \textbf{Input}:
    \begin{itemize}  
        \item \textit{"text input": "the tall green plant in the basket is standing near the woman in black top"}
        \item \textit{marked image}
    \end{itemize}
\end{itemize}

\begin{figure}[htb!]
\centering
\includegraphics[width=0.45\textwidth]{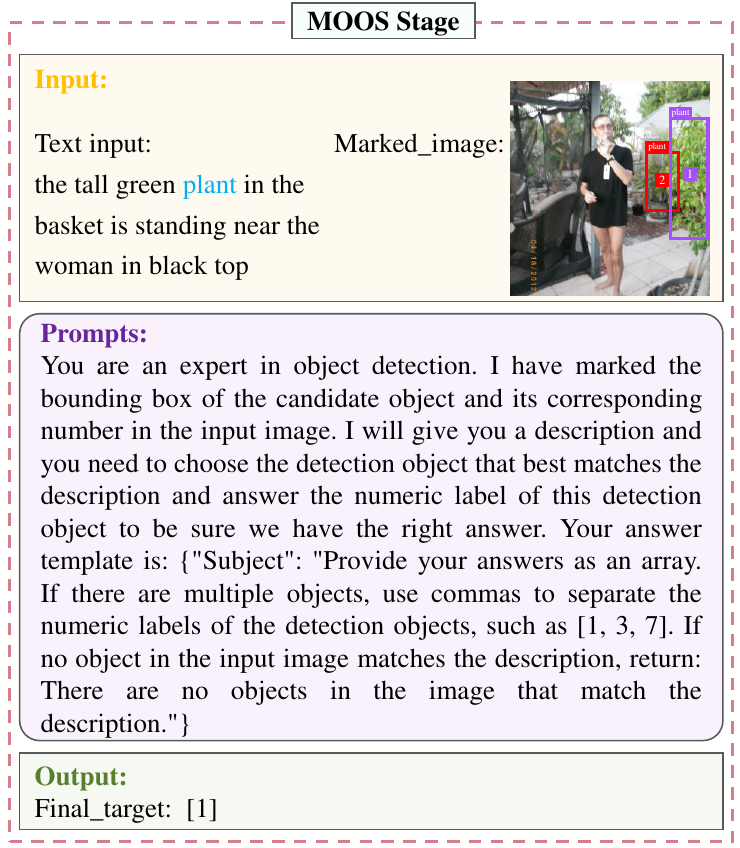}
\caption{Specific inputs, outputs, and corresponding prompts for the MOOS stage.}
\label{s3}
\end{figure}

In this stage, we use prompts, presented in Figure~\ref{s3}, to utilize MLLMs (GPT-4o and LLaVA-1.5) in aligning the intricate textual description with the optimal visual targets.
\begin{itemize}
    \item \textbf{Output}: Final reasoned targets generated by MLLMs, such as \textit{"final\_target": "[1]"}.
\end{itemize}

\section{More Qualitative Results}\label{qualitative_results}

Our MQADet paradigm achieves superior results compared with other state-of-the-art detectors. We exhibit the visualization results on GPT-4o and LLaVA-1.5 respectively.

\begin{figure*}[htb!]
\centering
\includegraphics[width=1.0\textwidth]{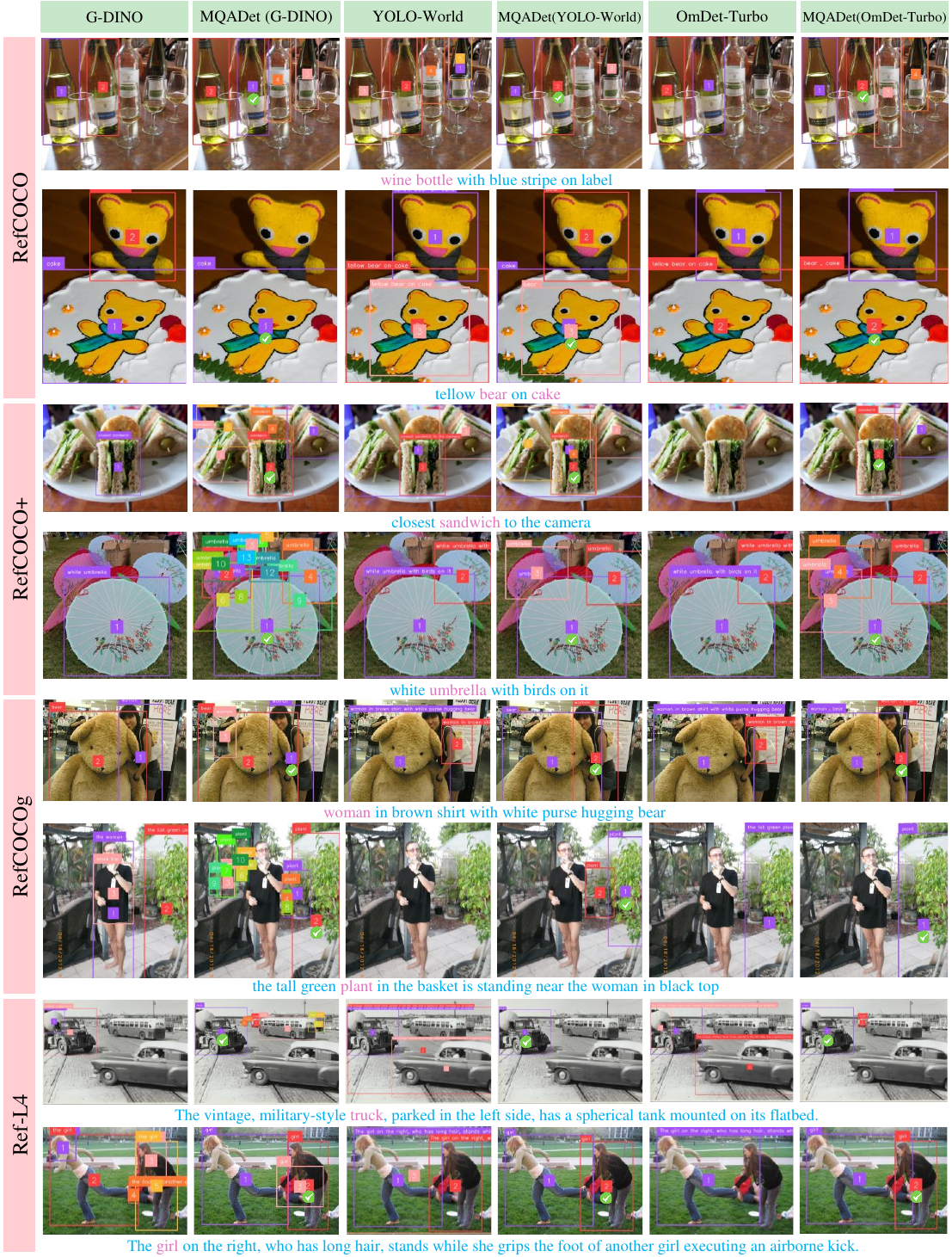}
\caption{Qualitative comparison between MQADet and three state-of-the-art OV detectors (Grounding DINO, YOLO-World, and OmDet-Turbo) on RefCOCO/+/g, and Ref-L4 datasets, with GPT-4o employed as the MLLM. \textcolor{mypink}{Pink} words denote the subjects identified from the user query.}
\label{gpt_visual}
\end{figure*}

\paragraph{Visualization on GPT-4o.} We compare the visualization results between MQADet and three state-of-the-art OV detectors (Grounding DINO, YOLO-World, and OmDet-Turbo) on RefCOCO, RefCOCO+, RefCOCOg, and Ref-L4 four benchmarks, with GPT-4o employed as the MLLM, as shown in Figure~\ref{gpt_visual}. Specifically, the first, third, and fifth columns display the predictions of Grounding DINO, YOLO-World, and OmDet-Turbo, respectively, while the remaining columns represent the results from our paradigm. The visualizations clearly illustrate that MQADet enables the detectors to focus on more categories by identified subject cues and leverages the vision-language reasoning capabilities of MLLM to better align fine-grained visual-textual information. These results stress the robust zero-shot detection capability of MQADet across different OV detection models on all benchmark datasets.

\begin{figure*}[htb!]
\centering
\includegraphics[width=1.0\textwidth]{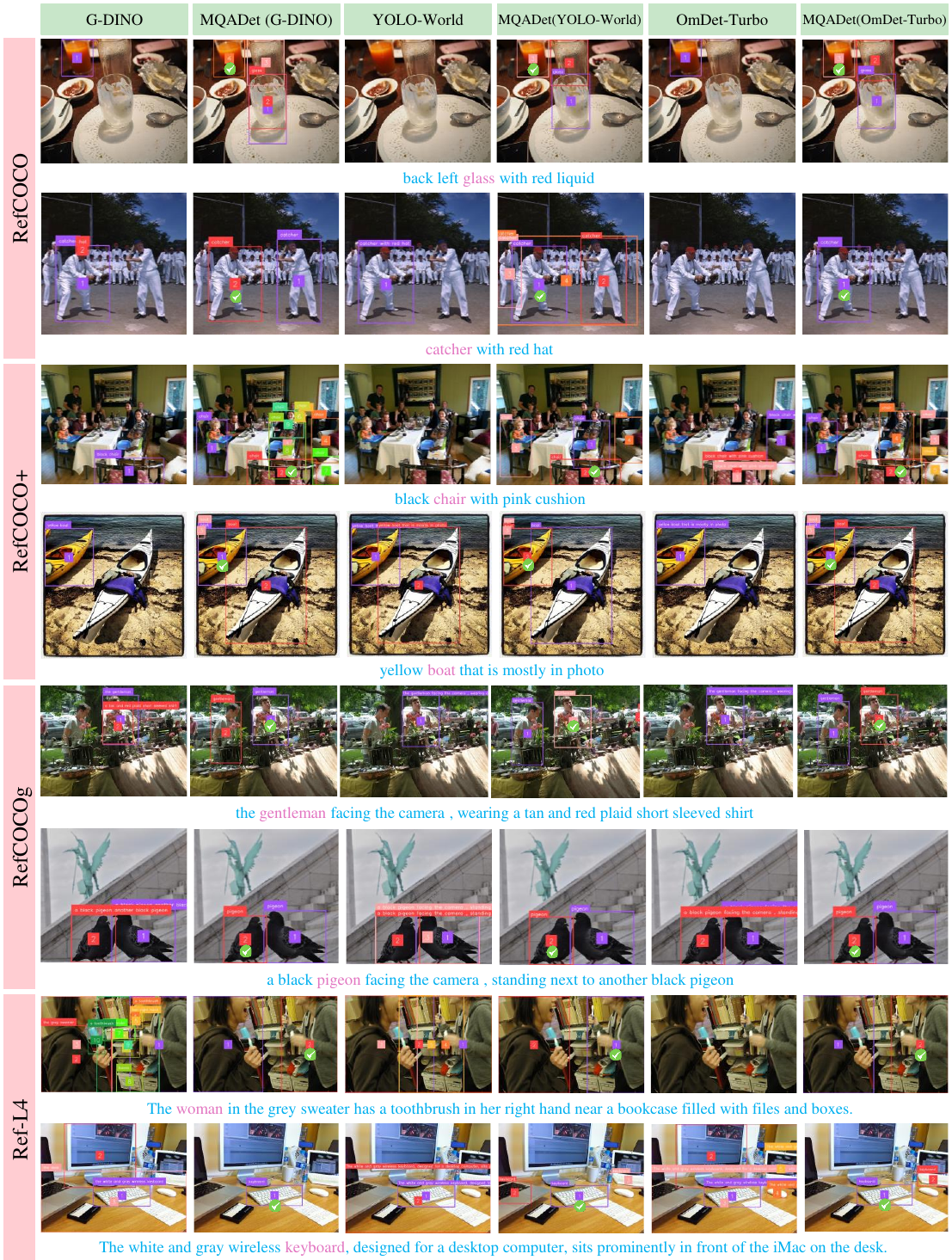}
\caption{Qualitative comparison between MQADet and three state-of-the-art OV detectors (Grounding DINO, YOLO-World, and OmDet-Turbo) on RefCOCO/+/g, and Ref-L4 datasets, with LLaVA-1.5 employed as the MLLM. \textcolor{mypink}{Pink} words denote the subjects identified from the user query.}
\label{llava_visual}
\end{figure*}

\paragraph{Visualization on LLaVA-1.5.} We then present the visualization results with LLaVA-1.5 employed as the MLLM on three detectors across all datasets to verify the transferability of our paradigm, as shown in Figure~\ref{llava_visual}. The visualizations demonstrate that, regardless of whether GPT or LLaVA is adopted as MLLMs, MQADet effectively bridges the logic gap between perception and reasoning, addressing the challenges of complex visual-textual misalignment and significantly improving the detection accuracy of existing detectors in open-vocabulary scenarios.

\bibliographystyle{named}
\bibliography{ijcai25}